\pdfoutput=1

\documentclass[11pt]{article}

\usepackage{acl}
\usepackage{comment}
\usepackage{times}
\usepackage{latexsym}
\usepackage{array}
\usepackage{booktabs}
\usepackage{caption}
\usepackage{tabularx}
\usepackage{ragged2e} 
\usepackage{natbib}
\usepackage{amssymb}
\usepackage{longtable}
\usepackage{lipsum} 
\usepackage{cuted} 
\usepackage[T1]{fontenc}

\usepackage[utf8]{inputenc}

\usepackage{microtype}

\usepackage{inconsolata}

\usepackage{graphicx}

\usepackage{rotating}

\usepackage{booktabs}  

\usepackage{amssymb}

\usepackage{kotex}

\definecolor{pred}{rgb}{0.7843, 0.0039, 0.3137}

\title{Towards LLM-Centric Multimodal Fusion: A Survey on Integration Strategies and Techniques}

\author{
 \textbf{Jisu An\textsuperscript{1}},
 \textbf{Junseok Lee\textsuperscript{1}},
 \textbf{Jeoungeun Lee\textsuperscript{2}},
 \textbf{Yongseok Son\textsuperscript{3}},
\\
 \textsuperscript{1}Seoul National University,
 \textsuperscript{2}University of California San Diego,
 \textsuperscript{3}Chung-Ang University,
\\
   \texttt{\{ajs7270,shawn159\}@snu.ac.kr} \\
    \texttt{jel166@ucsd.edu}, \texttt{sysganda@cau.ac.kr}
}

\begin{document}
\maketitle

\begin{abstract}
The rapid progress of Multimodal Large Language Models(MLLMs) has transformed the AI landscape. These models combine pre-trained LLMs with various modality encoders. This integration requires a systematic understanding of how different modalities connect to the language backbone.
Our survey presents an LLM-centric analysis of current approaches. We examine methods for transforming and aligning diverse modal inputs into the language embedding space. This addresses a significant gap in existing literature.
We propose a classification framework for MLLMs based on three key dimensions. First, we examine architectural strategies for modality integration. This includes both the specific integration mechanisms and the fusion level. Second, we categorize representation learning techniques as either joint or coordinate representations. Third, we analyze training paradigms, including training strategies and objective functions.
By examining 125 MLLMs developed between 2021 and 2025, we identify emerging patterns in the field. Our taxonomy provides researchers with a structured overview of current integration techniques. These insights aim to guide the development of more robust multimodal integration strategies for future models built on pre-trained foundations.
\end{abstract}
\section{Introduction}
\label{sec:introduction}
MLLMs represent a significant breakthrough in AI research. These models combine foundation LLMs with specialized encoders for different modalities including images, audio, and video~\cite{caffagni2024,wu2023a}.
Models that integrate multiple modalities achieve richer context comprehension by leveraging complementary information from different sources~\cite{shukangyin2023}.

\noindent\textbf{Prior studies:} Existing surveys provide valuable insights into multimodal LLMs. ~\citet{shukangyin2023} conducted a comprehensive review covering various aspects of multimodal LLMs including components, datasets, and training methodologies, but focused less on modality integration approaches. Similarly, ~\citet{song2025} attempted to categorize modality integration methods into multimodal convertor and multimodal perceiver, but this rigid classification creates confusion when architectural components serve multiple functional roles depending on implementation context. For example, an MLP(Multi-Layer Perceptron) layer might be used by ~\citet{liu2024} to project features into the LLM embedding space, while ~\citet{wang2024d} use it to reduce the number of image tokens.
This gap is particularly evident in architectural mechanisms, representation techniques, and training methodologies. A detailed comparison with prior surveys appears in Appendix~\ref{sec:appendix_related_survey}(Table ~\ref{tab:survey_comparison}).

\noindent\textbf{Key challenge:} Although the proliferation of MLLMs, a significant challenge persists in understanding the varied functional roles of common architectural components. Identical modules are often employed with differing contextual intentions across models. This forces researchers to invest considerable time in identifying the intended purpose of components, hindering the efficient design and adaptation of MLLMs.

\noindent\textbf{Contributions:} Our survey addresses this gap by analyzing 125 MLLM papers(2021-2025). We introduce a structured framework examining MLLMs through three key dimensions: 1) architectural strategies for modality fusion(including mechanisms such as Abstraction, Projection, Semantic Embedding, and Cross-attention layers, along with fusion levels like early, intermediate, or hybrid), 2) representation learning paradigms(joint or coordinate representation), and 3) training methodologies(training strategies and objective functions).


 


\newcolumntype{L}{>{\RaggedRight\arraybackslash}X} 
\begin{table*}[htbp]
\centering
 \tiny            

\begin{tabularx}{\textwidth}{@{} L L L L L @{}}
\toprule
\textbf{Model} & \textbf{Abstractor Layer} & \textbf{Projection Layer} & \textbf{Semantic Embedding Layer} & \textbf{Cross-attention Layer}  \\
\midrule
AnomalyGPT $\bigstar$\citep{gu2024} & - & Linear & Convolution & - \\
BLIP-2 $\bigstar$\citep{li2023e} & - & Linear & Q-former & -  \\
Cambrian-1 $\blacklozenge$\citep{tong2024} & - & - & Cross-attention (like Convolution) & - \\
CogAgent $\lozenge$ \citep{hong2024} & - & MLP & - & Within Model  \\
Flamingo $\blacklozenge$ \citep{alayrac2022} & Perceiver Resampler & - & - & Within Model \\
InstructBLIP \citep{dai2023} & - & Linear & Q-former & -  \\
Kosmos-1 \citep{huang2023} & Perceiver Resampler & MLP (last layer of ViT) & - & -  \\
LHRS-Bot \citep{muhtar2024} & - & - & Perceiver Resampler & - \\
LLaMA-Adapter V2 $\blacklozenge$ \citep{gao2023} & - & Linear & - & Within Model (self attention layer) \\
LLaVA \citep{liu2023} & - & Linear & - & -  \\
LLaVA-1.5 \citep{liu2024} & - & MLP & - & -  \\
MM1 \citep{mckinzie2024} & C-Abstractor & - & - & -  \\
mPLUG-Owl2 $\blacklozenge$ \citep{ye2024a} & Visual Abstractor & - & - & Modality Adaptive Module \\
mPLUG-Owl3 $\blacklozenge$ \citep{ye2024h} & - & Linear & - & Within Model (Hyper Attention Transformer block)  \\
PMC-VQA \citep{zhang2024e} & - & MLP / Transformer & - & -  \\
SEAL $\blacktriangle$\citep{sun2025} & Convolution & MLP & - & -  \\
SEED $\bigstar$\citep{ge2023} & - & Linear & Q-former & - \\
VideoChat \citep{k.-y.r.li2023} & Q-former & Linear & - & -  \\
VILA \citep{lin2024} & - & Linear / Transformer & - & -  \\
VisionLLM \citep{wang2023} & - & Transformer & - & -  \\
X-InstructBLIP \citep{panagopoulou2024} & - & Q-former + Linear & - & -  \\

\bottomrule
\end{tabularx}
\caption{Selected Comparisons of MLLMs. Key: Fusion Level is indicated by $\blacklozenge$ (Intermediate) or $\lozenge$ (Hybrid); other fusion instances are 'Early'. Representation is indicated by $\blacktriangle$ (Coordinate) or $\bigstar$ (Hybrid); other representation instances are 'Joint'.}
\label{tab:vl_models}
\end{table*}

Our primary contributions include:
\begin{itemize}
    \item We propose a novel cross-modality fusion mechanisms taxonomy that explains how the same architectural components can perform different contextual functions based on researchers' design intentions.
    \item We provide a cross-modality fusion mechanism table~\ref{tab:vl_models} that presents existing MLLM studies at a glance according to contextual fusion mechanisms and fusion levels.
    \item We identify emerging patterns and design principles from existing MLLM research to guide future multimodal system development.
\end{itemize}
This taxonomy provides researchers with a structured overview of current integration techniques for building MLLMs on pre-trained foundations.

The remainder of this paper is organized as follows.
Section~\ref{sec:background} provides essential background on foundational concepts.
Section~\ref{sec:architectural_strategies_for_modality_fusion} presents our classification taxonomy for architectural strategies.
Section~\ref{sec:representation} examines representation learning approaches.
Section~\ref{sec:data_and_training} investigates training methodologies.
Sections~\ref{sec:challenges_and_future_direction},~\ref{sec:conclusion}, and~\ref{sec:limitation} discuss future challenges, conclusions, and limitations respectively.

\section{Background}
\label{sec:background}
\paragraph{Multimodality}
Multimodality refers to combining different data types(text, images, audio, etc.) within a single AI system~\cite{ngiam2011,wu2023a,shukangyin2023,li2024b}. 
Humans understand the world through multiple senses, and similarly, AI systems perform better when using different input types together. 
When a model processes multiple modalities at once, it can capture information that single-modality models would miss~\cite{ngiam2011,karpathy2015}. 
For example, MLLMs combine visual and textual data to understand images in language terms~\cite{alayrac2022,pengwang2022,liu2023,li2023e,openai2024}. 
This integration enables better performance on tasks like image captioning and visual question answering by connecting language with visual perception~\cite{agrawal2017,driess2023,openai2024}.

\paragraph{Large Language Model}
LLMs form the foundation of MLLMs by providing strong language capabilities learned from massive text datasets~\cite{radford2018improving,radford2019language,brown2020}. 
These transformer-based models typically contain billions of parameters and excel at reasoning and few-shot learning~\cite{brown2020,chowdhery2023,wei2022,touvron2023,gunasekar2023,jiang2023a,team2025}. 
In most MLLMs, the pre-trained LLM remains frozen or lightly tuned to preserve its knowledge while reducing training costs~\cite{tsimpoukelli2021,li2023e,liu2023}. 
This approach leverages existing LLMs(like LLaMA or Vicuna) as reasoning engines without full finetuing, making multimodal system development more efficient~\cite{touvron2023,peng2023a,shukangyin2023}.

\paragraph{Feature Encoder}
MLLMs use specialized encoders to convert modality inputs into vector representations. 
For images, Vision Transformers(ViT) or convolutional networks extract visual features~\cite{krizhevsky2012,he2016,dosovitskiy2021}, while models like Wav2Vec~2.0 encode audio signals~\cite{schneider2019,baevski2020}. 
Many successful MLLMs use encoders with multimodal pre-training, such as CLIP, which aligns images and text in a shared embedding space~\cite{radford2021}. 
These pre-trained encoders provide features already aligned with language concepts, making it easier for the LLM to understand them. 
The modular design of MLLMs incorporates specialized encoders for different modalities while keeping them frozen to preserve their expertise~\cite{liu2023,pengwang2022,zhang2023a,k.-y.r.li2023}. 
Recent work on universal encoders, like ImageBind, aims to create common embedding spaces for multiple modalities, potentially simplifying future MLLM architectures~\cite{girdhar2023}.

\section{Architectural strategies for Modality fusion}
\label{sec:architectural_strategies_for_modality_fusion}

This section examines the key mechanisms for integrating non-text modalities with large language models. Understanding these integration mechanisms is crucial because researchers should able to select appropriate integration strategies based on their targeted purpose.

\subsection{Cross-modality fusion mechanisms}

\begin{figure}[t]
  \centering
  \includegraphics[width=\columnwidth]{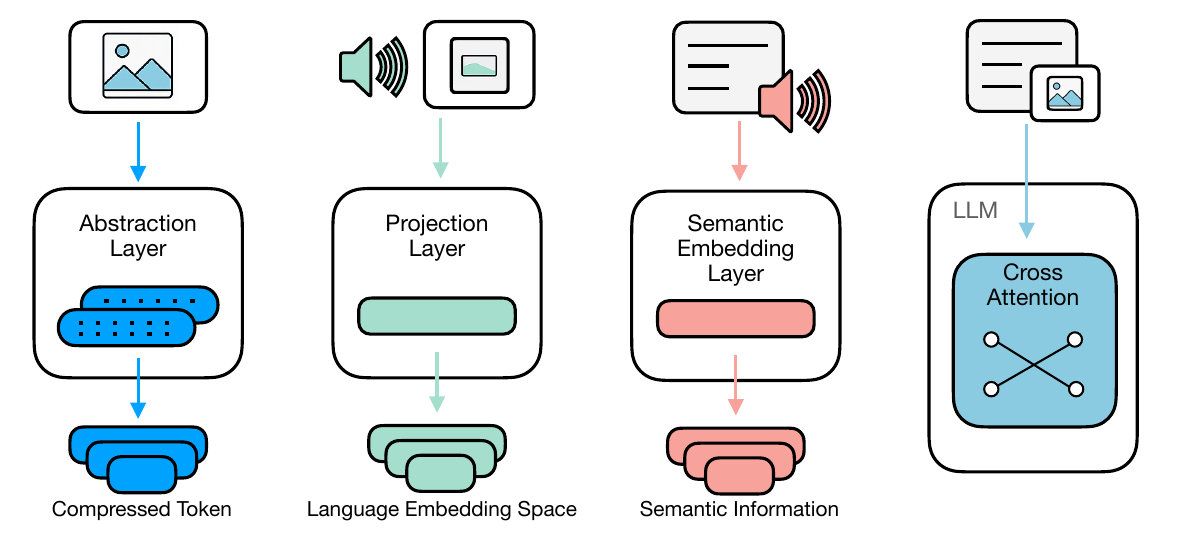}
  \caption{Proposed contextual fusion mechanisms}
  \label{fig:arch4fus}
\end{figure}

We propose a taxonomy of four contextual fusion mechanisms based on our analysis of recent multimodal LLM literature: \textbf{Projection}, \textbf{Abstraction}, \textbf{Semantic embedding} and \textbf{Cross-attention}. Figure~\ref{fig:arch4fus} illustrates the concept of the mechanisms. These mechanisms bridge the gap between different data types, allowing LLMs to process and generate responses based on diverse inputs.
These mechanisms can be implemented through various architectural components.
However, each component may serve multiple purpose depending on researcher's intention.
To address this confusion, this section examines how researchers implement these mechanisms to connect non-text features with language representations in the LLM's embedding space.

\subsubsection{Abstraction layer}

Abstraction layer is used to control the number of tokens from non-text modality features. When integrating non-text modalities with pre-trained LLMs, feature extraction using encoders is processed first. However, these extracted features undergo various processing to be used as input for LLMs. Abstraction layer acts as an information bottleneck\citep{mu2023} for extracted features, producing fewer or fixed-length tokens.

The abstraction process can mitigate several issues with extracted features. First, the number of features can vary depending on the data format. This variability in input feature count can increase the difficulty of model architecture design. Second, the number of features can be large. For example, a ViT encoder extracts features by dividing images into patches. Therefore, if the input image has high resolution, it results in numerous features. This large number of features can increase model size and computational costs during training and inference.

\citet{huang2023, laurencon2024, li2023a, xu2024a, yue2024, ye2024a} use Perceiver Resampler to achieve abstraction effects. Perceiver Resampler is a concept proposed in Flamingo\citep{alayrac2022}, with a Perceiver-based structure\citep{jaegle2021perceiver}. The main characteristic of Perceiver Resampler is using fixed-length learnable queries to perform cross-attention with input features, resulting in output tokens with a fixed length determined by the learnable queries.

\citet{zhang2023, tang2024, shao2024, somepalli2024} utilize Q-former as an abstraction layer. Q-former is a structure proposed in the BLIP-2\citep{li2023e}. Similar to Perceiver Resampler, Q-former uses learnable queries to fix the number of output tokens. The difference is that self-attention layers and cross-attention layers alternate. In self-attention layers, information exchange occurs between learnable queries, while in cross-attention layers, information exchange occurs between modality features and learnable queries. 

MM1\citep{mckinzie2024} uses C-Abstractor as an abstraction layer. C-Abstractor (Convolutional Abstractor) and D-Abstractor (Deformable attention-based Abstractor) are proposed by Honeybee\citep{cha2024honeybee}. They pointed out that Perceiver Resampler and Q-former based abstraction can inherently suffer from a risk of information loss. To address the limitations, the architectures are proposed. \citep{yu2024a, sun2025} used simple convolution layers for abstraction. Convolution layers are expected to capture spatial or temporal dependencies within features.

\subsubsection{Projection layer}

Projection layers map extracted features into the language embedding space, making them more comprehensible to the LLM. Using projection layers reduces the need to update LLM parameters for modality integration, which is economical and preserves the LLM's pretrained knowledge. This advantage has led to their widespread adoption in research.

 ~\citet{gao2023, k.-y.r.li2023, li2023c, lin2024, liu2023, su2023, zhang2023, ge2023} use linear layers as projection layers. The simple structure of linear layers allows for cross-modality integration effects with minimal resources. ~\citet{liu2024} mentions that linear layers may be insufficient for effective projection due to their simple structure.
~\citet{lai2024, liu2024, wang2024, hong2024, hu2024, xu2024, shao2024, jain2024, chen2024a} use more sophisticated MLPs for projection compared to linear layers. Notably, LLaVA-1.5~\citep{liu2024} improved performance by changing the projection layer from a linear layer to an MLP. ~\citet{lin2024, li2024e, zhang2024e} use transformer architecture which can serve enough capacity for projection. ~\citet{panagopoulou2024} use Q-former for instruction aware projection.

\subsubsection{Semantic embedding layer}

Beyond refining(by Abstraction Layer) or projecting extracted features, there are attempts to incorporate high-level information into features. Semantic embedding layers add high-level information to non-text features through various mechanisms.

Q-former is commonly used as a semantic embedding layer. The learnable queries used in Q-former can serve as instructions to extract semantic information from input features. ~\citet{li2023e, dai2023, hu2024, ren2024, ge2023, he2024a, chen2024a, qi2024, mittal2024} use Q-former as a layer for high-level semantics. InstructBLIP~\citep{dai2023} proposes a structure that explicitly includes text instructions as input to Q-former alongside learnable queries. ~\citet{mu2023, qi2024, chen2024a, hu2024, hu2023} use Q-former proposed in InstructBLIP to more explicitly incorporate high-level semantics. ~\citet{muhtar2024} use learnable queries in Perceiver Resampler to incorporate semantic information. ~\citet{gu2024} uses convolution layers for semantic information. ~\citet{tong2024} paper adds spatial inductive bias to vision features using learnable queries along with convolution-like cross-attention layers.

\subsubsection{Cross-attention layer}

A more explicit cross-modality integration method is utilizing cross-attention layers. Cross-attention layers allow LLMs to dynamically attend to non-text features. In cross-attention layers, query and key-value are generated from different modalities to enable inter-modality information exchange.
Since most modern LLMs have transformer structures, they already contain cross-attention layers. Modality integration can be induced by feeding non-text features into these internal cross-attention layers.

Flamingo~\citep{alayrac2022} induce modality integration by directly inputting non-text embeddings into internal cross-attention layers. ~\citet{gong2023, moor2023, li2023a, yue2024} follow similar structures based on Flamingo. CogVLM~\citep{wang2024} expands internal matrices for Query, Key, and Value for vision modality. CogAgent~\citep{hong2024} improves high-resolution image processing performance of CogVLM by feeding hidden states from CLIP-based vision encoder into internal cross-attention layers. mPLUG-Owl2~\citep{ye2024a} receives both vision and text inputs in the model's internal attention layers and uses projection matrices to generate modality-specific keys and values. mPLUG-Owl3~\citep{ye2024h} proposes Hyper Attention Transformer blocks that perform text-only self-attention and vision-text cross-attention at the same level before integration through adaptive gates. Besides using existing internal cross-attention layers, modality integration can be attempted by adding external cross-attention layers. LLaMA-VID~\citep{li2024d} adds separate cross-attention layers to compute cross-attention between text queries and vision embeddings.

\subsection{Fusion level}

\begin{figure*}[!t]
    \includegraphics[width=\textwidth]{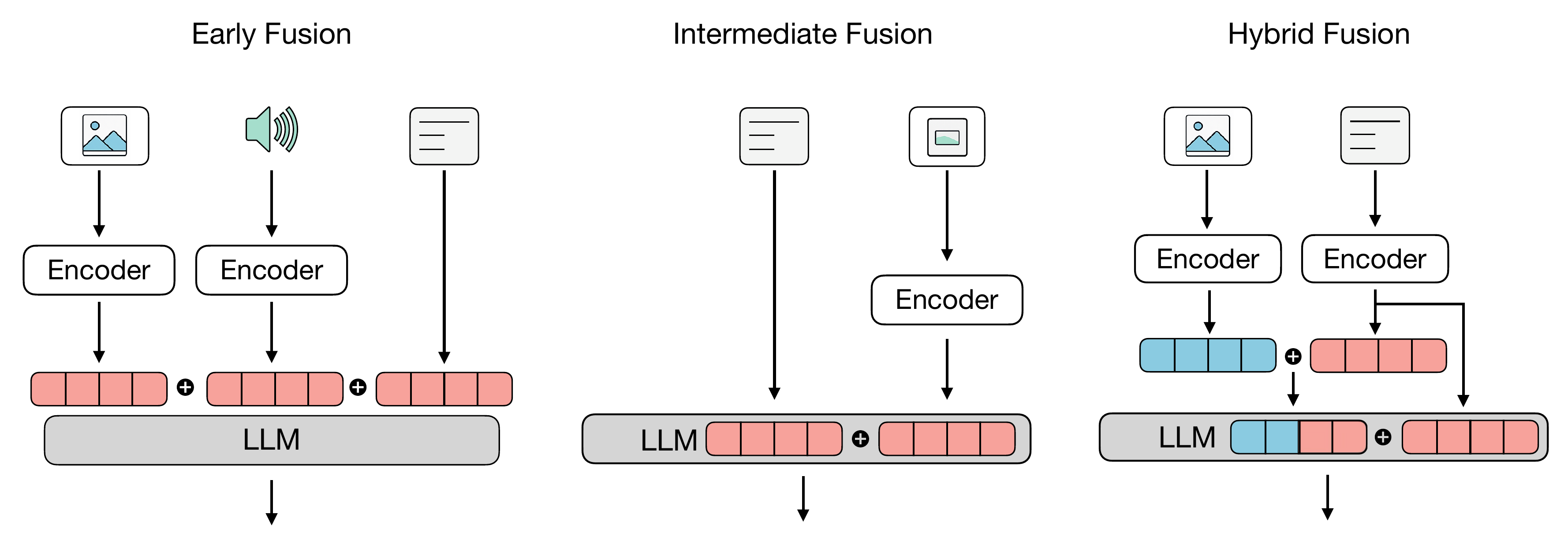}
    \caption{Proposed a taxonomy of LLM-centric fusion level(Early, Intermediate, and Hybrid)}
    \label{overview}
\end{figure*}

We propose a taxonomy of fusion strategies in MLLMs based on fusion level, i.e. the stage at which non-text modalities are integrated relative to the core LLM. As illustrated in Figure~\ref{overview}, we distinguish Early Fusion, Intermediate Fusion, and Hybrid Fusion.
These fusion levels are orthogonal to semantic fusion mechanisms(Projection, Abstraction, Cross-attention, Semantic Embedding): the same mechanism can be applied at different stages based on model design.

\subsubsection{Early Fusion}
Early Fusion merges non-text features before feeding them into the LLM, offering advantages in computational efficiency, model integrity preservation, and simplified integration. This approach processes and transforms modality embeddings into the LLM embedding space, reducing sequence length.

For instance, Kosmos-1~\citep{huang2023} merges patch tokens via an abstraction layer(Perceiver Resampler) to distill high-dimensional visual inputs into a compact representation. This early abstraction preserves salient information while reducing sequence length for computational efficiency. LLaVA~\citep{liu2023} projects CLIP embeddings through a linear projection layer into token embeddings compatible with the LLM, maintaining the pretrained backbone and minimizing integration overhead. BLIP-2~\citep{li2023e} employs a Q-Former semantic embedding module to extract focused visual queries, enriching language generation with high-level context and filtering out background noise to improve multimodal alignment. LLaMA-VID~\citep{li2024d} introduces an external cross-attention layer before LLM input to dynamically fuse projected vision tokens with text embeddings, enabling robust token alignment and dynamic feature selection while keeping the core LLM frozen.

\subsubsection{Intermediate Fusion}
Intermediate Fusion integrates non-text modalities within the LLM's transformer layers rather than before them. This approach allows for dynamic, layer-wise interaction between modalities throughout the processing pipeline.

For example, Flamingo~\citep{alayrac2022} uses an abstraction layer(Perceiver Resampler) to distill image features and gated cross-attention layers to fuse them with text. By placing cross-attention mid-layer, Flamingo dynamically queries visual details at each generation step, improving contextual grounding and reducing reliance on static embeddings. 
CogVLM~\citep{wang2024} maps vision embeddings via an MLP projection layer before using in-model cross-attention adapters. This preserves the frozen LLM backbone while enabling precise visual queries at each layer, balancing parameter efficiency and fine-grained cross-modal interaction.
Cambrian-1~\citep{tong2024} employs a convolutional semantic embedding layer via cross-attention, introducing spatial inductive bias directly into the LLM's layers. This enhances region-level alignment between visual features and text tokens, improving performance on spatial reasoning tasks.
LLaMA-Adapter V2~\citep{gao2023} attaches lightweight, zero-initialized cross-attention adapters to a frozen LLaMA, achieving multimodality with under 1M new parameters. Embedding adapters inside transformer layers allows seamless modality mixing with minimal overhead and preserves pretrained knowledge.
These intermediate fusion designs enable progressive, layer-wise modality merging, delivering precise grounding and task-specific reasoning with minimal model disturbance.

\subsubsection{Hybrid Fusion}
Hybrid Fusion combines the advantages of both Early and Intermediate fusion strategies, offering computational efficiency while maintaining fine-grained integration capabilities. This approach provides the preprocessing benefits of early fusion and the dynamic interaction benefits of intermediate fusion.

For example, CogAgent~\citep{hong2024} first projects vision features through an MLP before feeding them into the LLM, then uses cross-attention adapters to inject detailed interface elements, combining global layout and fine detail. This design leverages efficient early alignment and precise mid-layer fusion, making it ideal for vision understanding. ManipLLM~\citep{li2024e} injects cross-attention layers within transformer blocks to seamlessly merge modality tokens mid-stream. This semantic embedding approach preserves initial feature alignment while enabling dynamic, token-level integration—key for tasks that demand both global context and precise action grounding in real time. These hybrid systems combine the efficiency of early alignment with the precision of in-model fusion, offering both fast retrieval and deep, token-level integration for complex tasks.

\section{Representation of Multi-Modal Data}
\label{sec:representation}
The choice of representation learning on multimodal data—how heterogeneous inputs like text, images, audio, video, and sensor signals are integrated and processed—depends largely on the specific task at hand. 
 In this section, we categorize multimodal representations into three classes: \textbf{Joint}, \textbf{Coordinated}, and \textbf{Hybrid}, each suited to different objectives and requirements.

We discuss how Multimodal LLMs adopt these paradigms, outlining representative methods, their respective advantages and limitations.

\subsection{Joint Representation}
Joint Representation merges features from all modalities into one shared space. Different inputs (text, images, audio, video) become tokens in the same Transformer. The model fuses them via self-attention and cross-attention.

There are two main approaches. 
\noindent\textbf{Projection layer}: A projection or adapter maps each modality's features into the LLM's embedding space. These projected tokens join the text token sequence. Models like LLaVA~\citep{liu2023}, MiniGPT-4~\citep{zhu2024d}, LLaMA-Adapter~\citep{zhang2024b}, and others~\citep{liu2024,driess2023,pengwang2022,wang2024b,gao2023,chen2023,lai2024,zhang2024c,lin2024} use this method. This approach preserves the LLM's pre-trained structure and integrates modalities with minimal changes to the core model.

\noindent\textbf{Cross-attention layer}: The LLM is extended with cross-attention layers that let text queries attend to visual keys and values at selected transformer blocks. Flamingo~\citep{alayrac2022}, InstructBLIP~\citep{dai2023}, Qwen-VL~\citep{bai2023}, and similar models~\citep{wang2023,peng2023,zitkovich2023,ye2024b,zhu2024b,zhao2024} use cross-attention for detailed grounding. This approach enables fine-grained, token-level fusion by directing the model's attention to relevant modality features at each generation step. Additionally, several systems, such as MoVA and MMICL, combine cross-attention with adapter layers to achieve their specific goals. Because all modalities interact within a single network, joint methods excel at fine-grained visual question answering, image captioning, and multimodal dialogue. Nevertheless, they can struggle when a modality is missing at inference time and may be inefficient for retrieval-oriented tasks where independent modality embeddings are advantageous.

\subsection{Coordinated Representation}
Coordinated Representation uses separate encoders for each modality and contrastive learning to align their outputs. Matching pairs(e.g., image–text) are pulled together, while mismatched pairs are pushed apart. CLIP-style training is the canonical example: an image encoder and a text encoder are optimized so that embeddings of matching pairs converge while mismatched pairs diverge. Recent variants such as HACL~\citep{jiang2024} and DPE~\citep{zhang2024d} extend this idea.

This approach builds a shared embedding space that supports fast cross-modal retrieval and zero-shot transfer. Encoders stay modular: they can be swapped or updated without modifying the LLM. It also handles missing modalities gracefully and scales to new data.

Choose coordinated representation when retrieval efficiency, encoder flexibility, and robustness are priorities. However, without a fusion module, it cannot capture fine-grained token-level grounding or complex reasoning. In those cases, a joint representation may be more effective.

\subsection{Hybrid Representation}
Hybrid Representation combines coordinated alignment and joint fusion in two steps. First, separate encoders extract modality embeddings and align them with a contrastive loss(e.g., CLIP-style). Second, a fusion module—such as a cross-attention Transformer block or Q-Former—merges these embeddings into a unified representation. Models like BLIP-2~\citep{li2022,li2023e}, MoVA~\citep{zong2024}, and Video-LLaMA~\citep{zhang2023} follow this pattern.

This design keeps encoders modular for fast cross-modal retrieval and zero-shot transfer, while the fusion step enables fine-grained, token-level grounding for detailed reasoning. Choose hybrid representation when you need both efficient search and precise multimodal integration. If you only need deep fusion without retrieval, a joint representation may suffice; if you only need fast similarity search, a coordinated representation is simpler.

Selecting a representation depends on your data, inference setup, and task needs. Joint representation is best for deep feature fusion and fine-grained reasoning. Coordinated representation excels at fast retrieval and modular encoder updates. Hybrid representation offers both search efficiency and detailed grounding. Future MLLMs may dynamically switch or combine these modes based on input modality and task context. As models handle more modalities(audio, video, 3D, robotic data) under real-time demands, representation methods must stay efficient and scalable.
    
\section{Training Strategies and Objectives }
\label{sec:data_and_training}
The training process for modality integration can be categorized as \textbf{Single-stage training}, \textbf{Two-stage training}, or \textbf{Multi-stage training} based on the number of training phases.

\subsection{Single-stage training}
~\citet{li2022, tsimpoukelli2021, dai2023, su2023, li2023a} uses a single training step to induce modality integration. Integrating modalities into LLMs requires enormous cross-modal datasets, and captioning datasets are commonly used for this purpose. However, simple pairwise datasets like captioning datasets may not be sufficient for modality integration~\citep{zhu2024d}. Therefore, mixing various datasets for training is also utilized ~\citep{zitkovich2023, driess2023, lai2024}.

\subsection{Two-stage training}
To address the limitations of single-stage training, many approaches use multiple stages~\citep{liu2024, li2023e, zhu2024d, zhang2023, peng2023, k.-y.r.li2023, wang2024, gao2023}. The 2-stage approach typically consists of an alignment stage and an instruction tuning stage. In the alignment stage, a large amount of pairwise captioning data is used to train alignment modules such as projection layers. To enhance instruction following capabilities that might be lacking in captioning dataset training, the second stage uses a smaller amount of sophisticated instruction datasets for instruction tuning.

The 2-stage approach offers several benefits. First, it allows the use of both large but simple datasets and small but sophisticated datasets. Second, it mitigates catastrophic forgetting. When integrating new modalities into LLMs, it's important to effectively utilize the LLM's pre-trained knowledge. However, catastrophic forgetting can occur during the process of updating LLM parameters. In most 2-stage approaches, alignment is performed while keeping the LLM frozen. This helps mitigate the catastrophic forgetting problem compared to techniques that update model parameters throughout the entire training process.

\subsection{Multi-stage training}
Some studies attempt multi-stage training by more precisely controlling the dataset and target components. Multi-stage training is used to change target components by stage~\citep{bai2023}, vary the datasets used by stage~\citep{liu2025, zhang2023a, wang2024c, chen2023c}, or expand datasets stage by stage from a curriculum learning perspective~\citep{li2023c, gong2024, muhtar2024, chen2024c, zhu2024b}.


\subsection{Training Objectives}
To integrate non-text modalities, MLLMs are trained with a language modeling loss. This loss measures how well the model predicts the next text token given previous tokens and multimodal features. It is typically implemented as cross-entropy or negative log-likelihood over the token distribution. By using this objective, the MLLM learns both fluent language generation and alignment with non-text inputs.

Most MLLM research uses this LM objective, though many works also combine it with various losses for better modality alignment and grounding. Some two-stage approaches use different training objectives between the first and second stages. ~\citet{li2022, mckinzie2024, deshmukh2023, li2024a} use contrastive loss in the first stage to align different modalities. During this process, modality encoders are trained to produce aligned embeddings. In the second stage, the language modeling loss is used to integrate these aligned representations with the LLM.

Reconstruction loss is also used for modality integration~\citep{ge2023, yang2024, pan2024, zhu2024}. These studies aim to reconstruct non-text modalities as the output of modality integration. In the cross-modality reconstruction process using LLMs, the LLM creates conditions for reconstruction. The goal is to generate appropriate conditions so that outputs can be reconstructed by understanding the context contained in cross-modality data. Additionally, in the vision-language field, DICE loss is sometimes used for segmentation performance~\citep{lai2024, gu2024, zhang2024a}.

\section{Challenges and Future direction}
\label{sec:challenges_and_future_direction}
In our survey of MLLMs, we introduce several key research challenges and promising future research directions. MLLMs face several technical limitations including hallucination problems~\citep{xu2024b}, inconsistencies between modalities, and vulnerability to adversarial attacks~\cite{shayegani2023,zhao2023a,jeong2024}. Recent work has highlighted these issues, showing how carefully crafted inputs can exploit weaknesses in multimodal processing.

Based on our analysis, we identify several promising research directions. First, incorporating Retrieval-Augmented Generation(RAG) into multimodal systems shows potential for improving factual accuracy. Second, developing multimodal agents with enhanced reasoning capabilities represents an important frontier. Third, current MLLMs lack persistent memory mechanisms beyond context windows—an essential capability for more general intelligence that would allow models to leverage previous interactions without explicit context injection. Finally, while reasoning-focused LLMs~\citep{openai2024a, deepseek-ai2025,yang2025} have made significant progress, their multimodal counterparts remain underdeveloped in open-source research, presenting a critical opportunity. Addressing these challenges will advance the development of more robust and capable multimodal systems.







\section{Conclusion}
\label{sec:conclusion}

Multimodal LLMs extend language models by integrating visual, audio, and other modalities through abstraction, projection, semantic embedding and cross-attention on different fusion levels. We introduce a clear taxonomy based on three dimensions: Architectural strategies for Modality fusion, representation paradigms(joint, coordinated, hybrid), and training paradigms(single-stage, two-stage, multi-stage). By reviewing 125 MLLMs, we reveal the intentions and design philosophies of researchers, showing how they approached challenges in effectively and efficiency. We examine ongoing challenges in evaluation, multimodal hallucination, and persistent memory, while highlighting promising research directions such as retrieval-augmented generation and reasoning-focused MLLMs. This survey provides future researchers with insights into the evolution of design choices and research directions, offering a roadmap that builds upon previous work to guide the development of more robust and efficient multimodal language models.

\section{Limiation}
\label{sec:limitation}

This survey analyzes MLLM architectures and methods qualitatively, but has several limitations.
First, cross-modal evaluation remains difficult due to the lack of standardized metrics across different modalities. The absence of common benchmarks hinders effective comparison between models, obscuring performance differences. 
Therefore, we do not provide new experiments or direct comparisons between models. We rely on findings reported in original papers.
Second, many papers implementing specific integration mechanisms do not clearly state their design motivations—this is particularly problematic since the same architectural components often serve different functional purposes across studies. 
Thus, we focused on models with clear descriptions of their integration mechanisms(projection layers, cross-attention, etc.) and design choices. Models with unclear architectural details are excluded from our analysis tables.
Third, we examine MLLMs with direct integration between modalities and the LLM backbone. We do not cover indirect integration approaches, such as LLMs connecting to other modalities through intermediate models(like diffusion models) with shared loss functions.
Lastly, most surveyed MLLMs use language modeling loss for training. The lack of models using different loss functions limits our ability to compare diverse training objectives. This reflects current research trends rather than an intentional exclusion.

\bibliography{references}

\appendix
\section{related survey}
\label{sec:appendix_related_survey}

\begin{table*}[b] 
  \centering
  \caption{Comparison with existing MLLM surveys based on our proposed classification dimensions. Coverage is indicated by symbols (\checkmark: Covered, $\times$: Not Covered, $\sim$: Partial/Different). Columns represent cited surveys (ordered based on new data, mostly chronological) and the current work (Ours).} 
  \label{tab:survey_comparison} 
  \resizebox{\textwidth}{!}{
  \begin{tabular}{@{}lccccccccc|c@{}}
    \toprule
    Dimension / Survey &
    \rotatebox{90}{\citet{shukangyin2023}} &
    \rotatebox{90}{\citet{wu2023a}} &
    \rotatebox{90}{\citet{caffagni2024}} &
    \rotatebox{90}{\citet{jin2024b}} &
    \rotatebox{90}{\citet{li2024b}} &
    \rotatebox{90}{\citet{zhang2024f}} &
    \rotatebox{90}{\citet{han2025}} &
    \rotatebox{90}{\citet{jiang2025a}} &
    \rotatebox{90}{\citet{song2025}} &
    \rotatebox{90}{\textbf{Ours}   } \\
    \midrule

    (1) Semantic Int. Mech. & $\sim$\tiny{(Fig 2)} & $\times$ & $\sim$\tiny{(Sec 2.3)} & $\sim$\tiny{(Sec 2.2)} & $\times$ & $\sim$\tiny{(Sec 2.2)} & $\times$ & $\times$ & $\sim$\tiny{(Sec 3,4)} & \checkmark \\
    (2) Fusion Level (LLM-relative) & $\sim$\tiny{(Diff.fusion levels)} & $\times$ & $\times$ & $\times$ & $\sim$\tiny{(Diff. fusion levels)} & $\times$ & $\sim$\tiny{(no classification applied)} & $\times$ & $\times$ & \checkmark \\
    (3) Data \& Training Paradigm & $\times$ & $\times$ & $\times$ & $\times$ & $\times$ & $\times$ & $\times$ & $\times$ & $\times$ & \checkmark\\
    (4) Rep. Learning Approach & $\times$ & $\times$ & $\times$ & $\times$ & $\times$ & $\times$ & $\times$ & $\times$ & $\times$ & \checkmark \\
    \bottomrule
  \end{tabular}%
  } 
\end{table*}

This appendix compares nine representative MLLM surveys across four key dimensions: (1) Semantic Fusion Mechanisms (Projection, Abstraction, Cross-attention, Semantic Embedding), (2) LLM-relative Fusion Levels (Early, Intermediate, Hybrid), (3) Data \& Training Paradigms (Contrastive, Generative, Hybrid), and (4) Representation Learning Approaches (Joint vs. Coordinated).
None of the existing surveys covers all four dimensions cohesively. For example, \citet{shukangyin2023} introduces token vs. feature fusion but lacks semantic mechanism analysis and representation taxonomy. \citet{li2024b} proposes data- and feature-level fusion but does not classify LLM-centric fusion levels or representation approaches. \citet{caffagni2024} reviews two-stage training but omits detailed loss function comparisons and representation learning frameworks. \citet{jiang2025a} offers a unified modeling perspective but does not systematically examine fusion levels or training paradigms. Surveys by \citet{wu2023a}, \citet{jin2024b}, \citet{zhang2024f}, and \citet{song2025} each cover select aspects but do not provide the integrated view across all four dimensions. Our work addresses this gap by offering a unified framework along these dimensions (see Table \ref{tab:survey_comparison}).

\label{sec:appendix_related_survey}

 \begin{figure*}[t]
   \centering
   \includegraphics[width=\textwidth]{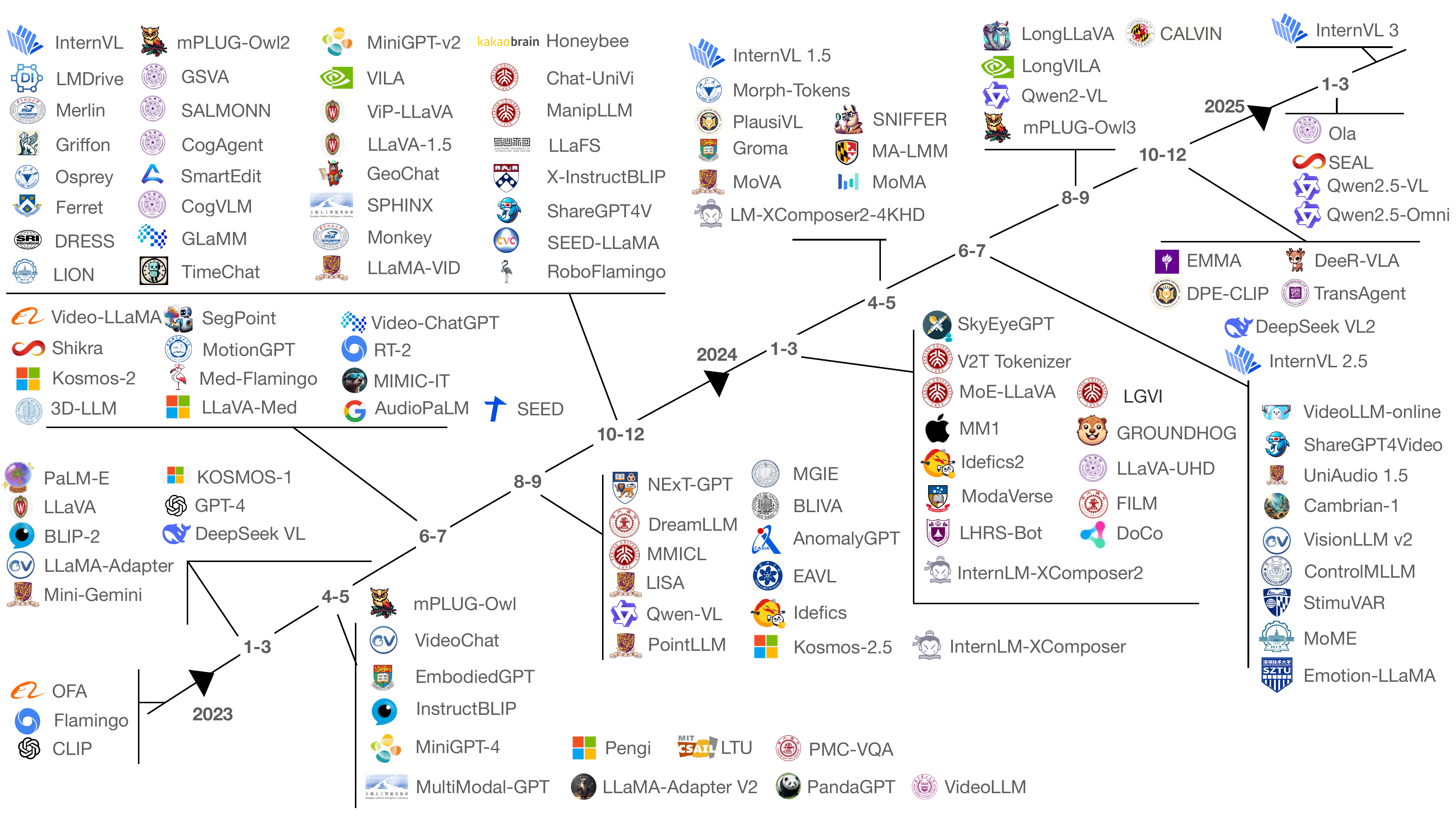}
   \caption{Timeline of major developments in multimodal architectures}
   \label{fig:timeline}
 \end{figure*}


\newcolumntype{L}{>{\RaggedRight\arraybackslash}X} 
\begin{table*}[htbp]
\centering
 \tiny            

\begin{tabularx}{\textwidth}{@{} L L L L L c c@{}}
\toprule
\textbf{Model} & \textbf{Abstractor Layer} & \textbf{Projection Layer} & \textbf{Semantic Embedding Layer} & \textbf{Cross-attention Layer}  & \textbf{Fusion level}  & \textbf{Representation}  \\
\midrule
EmbodiedGPT\citep{mu2023} & - & Linear & Q-former & - & Early & Joint \\
Flamingo\citep{alayrac2022} & Perceiver Resampler & - & - & Within Model & Inter & Joint \\
Kosmos-1\citep{huang2023} & Perceiver Resampler & MLP(last layer of ViT) & - & - & Early & Joint \\
Idefics2\citep{laurencon2024} & Perceiver Resampler & MLP & - & - & Early & Joint \\
RoboFlamingo\citep{li2023a} & Perceiver Resampler & - & Cross-attention(within Model) & - & Inter & Joint \\
LLaVA-UHD\citep{xu2024a} & Perceiver Resampler & - & - & - & Early & Joint \\
DeeR-VLA\citep{yue2024} & Perceiver Resampler & - & - & - & Inter & Joint \\
BLIP-2\citep{li2023e} & - & Linear & Q-former & - & Early & Hybrid \\
MM1\citep{mckinzie2024} & C-Abstractor & - & - & - & Early & Joint \\
LLaVA-1.5\citep{liu2024} & - & MLP & - & - & Early & Joint \\
InstructBLIP\citep{dai2023} & - & Linear & Q-former & - & Early & Joint \\
CogVLM\citep{wang2024} & - & MLP & - & within Model & Inter & Joint \\
CogAgent\citep{hong2024} & - & MLP & - & within Model & Hybrid & Joint \\
mPLUG-Owl2\citep{ye2024a} & VIsual Abstractor & - & - & Modality Adaptive Module & Inter & Joint \\
mPLUG-Owl3\citep{ye2024h} & - & Linear & - & within Model(Hyper Attention Transformer block) & Inter & Joint \\
LLaMA-VID\citep{li2024d} & - & Linear & - & external layer & Early & Joint \\
Video-LLaMA\citep{zhang2023} & Q-former & Linear & - & - & Early & Joint \\
SALMONN\citep{tang2024} & Q-former & - & - & - & Early & Joint \\
LMDrive\citep{shao2024} & Q-former & MLP & - & - & Early & Joint \\
CALVIN\citep{somepalli2024} & Q-former & Linear & - & - & Early & Joint \\
Merlin\citep{yu2024a} & Convolution & - & - & - & Early & Joint \\
SEAL\citep{sun2025} & Convolution & MLP & - & - & Early & Coord \\
LLaMA-Adapter V2\citep{gao2023} & - & Linear & - & within Model(self attention layer) & Inter & Joint \\
VideoChat\citep{k.-y.r.li2023} & Q-former & Linear & - & - & Early & Joint \\
LLaVA-Med\citep{li2023c} & - & Linear & - & - & Early & Joint \\
VILA\citep{lin2024} & - & Linear / Transformer & - & - & Early & Joint \\
PandaGPT\citep{su2023} & - & Linear & - & - & Early & Joint \\
SEED\citep{ge2023} & - & Linear & Q-former & - & Early & Hybrid \\
MoE-LLaVA\citep{lin2024a} & - & Linear & Q-former & - & Early & Joint \\
VisionLLM\citep{wang2023} & - & Transformer & - & - & Early & Joint \\
ManipLLM\citep{li2024e} & - & Transformer & - & within Model(self attention layer) & Hybrid & Joint \\
LongVILA\citep{chen2024c} & - & Linear / Transformer & - & - & Early & Joint \\
PMC-VQA\citep{zhang2024e} & - & MLP / Transformer & - & - & Early & Joint \\
X-InstructBLIP\citep{panagopoulou2024} & - & Q-former + Linear & - & - & Early & Joint \\
VisionLLM v2\citep{wu2024b} & - & Q-former + Linear & - & - & Early & Joint \\
MiniGPT-4\citep{zhu2024d} & - & Linear & - & - & Early & Joint \\
TimeChat\citep{ren2024} & - & Linear & Q-former & - & Early & Joint \\
MA-LMM\citep{he2024a} & - & - & Q-former & - & Early & Joint \\
LION\citep{chen2024a} & - & MLP & Q-former & - & Early & Joint \\
Sniffer\citep{qi2024} & - & - & Q-former & - & Early & Joint \\
PlausiVL\citep{mittal2024} & - & Linear & Q-former & - & Early & Hybrid \\
VideoLLM-online\citep{chen2024a} & - & MLP & - & - & Early & Joint \\ 
Beyond Text\citep{zhu2024} & - & Linear & - & - & Early & Joint \\
RSGPT\citep{hu2023} & - & - & Q-former & - & Early & Joint \\
LHRS-Bot\citep{muhtar2024} & - & - & Perceiver Resampler & - & Early & Joint \\
AnomalyGPT\citep{gu2024} & - & Linear & Convolution & - & Early & Hybrid \\
Cambrian-1\citep{tong2024} & - & - & Cross-attention(like Convolution) & - & Inter & Joint \\
MultiModal-GPT\citep{gong2023} & Perceiver Resampler & - & - & Within Model & Inter & Joint \\
Med-Flamingo\citep{moor2023} & Perceiver Resampler & - & - & Within Model & Inter & Joint \\
RoboFlamingo\citep{li2023a} & Perceiver Resampler & - & - & Within Model & Inter & Joint \\ 

\bottomrule
\end{tabularx}
\caption{Comparison of 50 Multi-Modal Architectures.} 
\label{tab:multimodal_architectures}
\end{table*}

\end{document}